\documentclass[a4paper,10pt]{article}
\usepackage{amsmath,amstext,amsgen,amsbsy,amsopn,amsfonts,amssymb}
\usepackage{etex}
\usepackage{graphics}
\usepackage[pdftex]{graphicx}
\usepackage{epsfig}
\usepackage{hyperref}
\usepackage{amsthm}
\usepackage{float}
\usepackage{bm}
\usepackage{color}
\usepackage{algorithmic}
\usepackage{algorithm}
\usepackage{quoting}
\usepackage{flushend}
\usepackage{balance}
\usepackage{multicol}
\usepackage[resetlabels]{multibib}

\newtheorem{theorem}{Theorem}[section]

\usepackage{booktabs}
\usepackage{esvect}
\usepackage{multirow}
\usepackage{array}
\usepackage{arydshln}
\usepackage{xcolor}
\usepackage{hyperref}

\providecommand{\keywords}[1]{\textbf{\textit{Index terms---}} #1}

\begin{document}
\title{\textbf{$DC^2$: A Divide-and-conquer Algorithm for Large-scale Kernel Learning with Application to Clustering}}

\author{
Ke Alexander Wang$^{\dag*}$, Xinran Bian$^{\ddag}$\footnote{Contributed equally.}, Pan Liu$^{\P}$, 
Donghui Yan$^{\$}$
\vspace{0.1in}
\\$^{\dag}$Cornell University, Ithaca, NY\vspace{0.05in}\\
$^{\ddag}$Shanghai Jiao Tong University, China\vspace{0.05in}\\
$^\P$Zhejiang University, China\vspace{0.05in}\\
$^{\$}$University of Massachusetts Dartmouth, MA\vspace{0.05in}
}

\date{\today}
\maketitle

\begin{abstract}
\noindent
Divide-and-conquer is a general strategy to deal with large scale problems. It is typically applied to generate ensemble instances, which 
potentially limits the problem size it can handle. Additionally, the data are often divided by random sampling which may be suboptimal. To 
address these concerns, we propose the $DC^2$ algorithm. Instead of ensemble instances, we produce structure-preserving signature 
pieces to be assembled and conquered. $DC^2$ achieves the efficiency of sampling-based large scale kernel methods while enabling
parallel multicore or clustered computation. The data partition and subsequent compression are unified by recursive random projections. 
Empirically dividing the data by random projections induces smaller mean squared approximation errors than conventional random 
sampling. The power of $DC^2$ is demonstrated by our clustering algorithm rpfCluster\textsuperscript{+}, which is as accurate 
as some fastest approximate spectral clustering algorithms while maintaining a running time close to that of K-means clustering. Analysis on 
$DC^2$ when applied to spectral clustering shows that the loss in clustering accuracy 
due to data division and reduction is upper bounded by the data approximation error which would vanish with recursive random projections. 
Due to its easy implementation and flexibility, we expect $DC^2$ to be applicable to general large scale learning problems.
\end{abstract}


\section{Introduction}
\label{section:introduction}
Kernel learning is an important problem in machine learning that lies at the core of kernel methods~\cite{ScholkopfSmola2001,HofmannScholkopfSmola2008}. 
For example, support vector machines \cite{CortesVapnik1995}, and various kernelized methods such as kernel PCA \cite{Scholkopf1998}, 
kernel ridge regression \cite{kernelRidge1998}, kernel ICA \cite{BachJordan2003}, kernel CCA \cite{FukumizuBachGretton2007}, kernel 
k-means \cite{DhillonGK2004} etc all require the learning of a kernel. Additionally, kernels have been used in semi-supervised learning, for example, \cite{BelkinMN2004} 
uses the graph Laplacian. Furthermore, all spectral clustering algorithms \cite{Ncut, NgJordan2002, Luxburg2007, YanHuangJordan2009tech} 
involve the learning of a similarity kernel.
\\
\\
There are several attractive properties with kernels. It allows the embedding of potentially unstructured data to a space 
suitable for learning and inference, and often effectively overcomes the {\it curse of dimensionality} arising from high dimensional data. 
However, as the kernel requires the computing of pairwise similarity for all points, the computational complexity for learning the kernel
is very high ($O(n^2)$ for a naive implementation on $n$ points). Indeed many spectral clustering algorithms, e.g., \cite{Ncut, NgJordan2002}, 
have a computational complexity of $O(n^3)$. A number of algorithms have been proposed to speed up the computation. For example, 
the Nystr\"om algorithm \cite{NystromSpectral, NystromApproximation}, and some data-dependent sampling algorithms \cite{YanHuangJordan2009tech, Chen2011LargeSS}. 
While such algorithms generally work remarkably well, they do not take advantage of todays' widely available multicore or 
clustered computing infrastructure. We propose a divide-and-conquer approach to split a large-scale kernel learning problem such that 
subtasks can run in parallel on multicore or clustered computers.    
\\
\\  
{\it Divide-and-conquer} is a popular strategy to deal with large or complex
problems when the subproblems are easier to solve or faster to compute.
Divide-and-conquer works by dividing a problem into smaller subproblems,
working on the subproblems, and then aggregating results from subproblems into
the solution to the original problem. Classic divide-and-conquer algorithms
include Quicksort \cite{Hoare1962}, the Karatsuba algorithm for multiplying
large numbers \cite{KaratsubaOfman1962}, fast Fourier transform 
\cite{FFT} etc. For large scale computation, there are generally two possible
ways to implement divide-and-conquer. One is to divide the algorithm into
parallel components and conquer on each. However, this is algorithm dependent
since it requires tailoring the divide-and-conquer to the algorithm's implementation details.
Another approach, which is popular and easy to implement, is to partition the data such that
each partition forms a subproblem.  As the subproblems use different partitions of
the data, they can run independently. This allows one to take advantage of the
multicore or clustered computing infrastructure where subproblems are computed
in parallel to  speed up the overall computation. If the target algorithm
has a super-linear computational complexity and the aggregation of subproblems
is ``easy'', one can still achieve remarkable speedup by solving the
subproblems {\it sequentially}, especially when the
original computation takes large computing resources. Dividing the data and
then conquering each component also makes it possible to tackle a larger problem than forming an
ensemble instance on each data partition, as different partitions can be used
to learn the target (e.g., data representation) on disjoint supports.  In
contrast, directly ensembling on the similarity matrix would severely limit
scalability as each partition now has to learn the full similarity matrix which
is not desirable due to its $O(n^2)$ complexity.
\\
\\
The most straightforward way to divide the data is by random sampling. This is particularly easy to implement and suitable for 
ensemble-based algorithms. However, random sampling does not account for
structure in the data and may be suboptimal. As a remedy, we propose to use
random projections to divide the data. 
Our method gives more accurate and more robust results by taking into account
the geometry of the data.  Given the high
computational complexity of kernel learning, a further representation
compression over each partition is often necessary. Representation compression
obtains a structure-preserving signature of the data which can be used in place
of the full dataset to speed up computations. We will use recursive random projections
\cite{DasguptaFreund2008, rpForestsArXiv2018} to produce a compressed signature
for each partition. The idea of recursive random projections has been
successfully applied in fast approximate spectral clustering
\cite{YanHuangJordan2009tech}, computing over distributed data
\cite{distSpectArXiv2019, distStatArXiv2019}, and other procedures. 
Since our approach is a divide-and-conquer method with a representation compression, we refer to it as
{\it divide-compress-and-conquer}, or $DC^2$ in short. 
\\
\\
Our main contributions are as follows. First, we propose a geometry-aware divide-compress-and-conquer algorithm for large scale 
kernel learning. It unifies the division of the data and the subsequent representation compression over each partition through recursive 
random projections. It incurs negligible approximation error in the resulting kernel, and a vanishing loss in accuracy when applied to 
clustering. As our approach is based on data partition, it allows parallel computation of subtasks on each data partition with multicore 
or clustered computing. Our proposed algorithm is easy to implement and readily applies to high dimensional data without the potential 
curse of dimensionality. Beyond kernel learning, it is immediately applicable to general large scale learning and inference problems. 
\\
\\
The remainder of this paper is organized as follows. In Section \ref{section:method}, we give a detailed description of the 
{\it $DC^2$} algorithm and its application to a kernel-based clustering algorithm. This is followed by 
a theoretical analysis on the approximation error of {\it $DC^2$} and the resulting loss in clustering accuracy. Related work are discussed 
in Section \ref{section:related}. In Section~\ref{section:evaluation}, we empirically evaluate the approximation error by 
recursive random projections in {\it $DC^2$} and we compare a {\it $DC^2$}-enabled clustering algorithm to its competitors. 
Finally, we conclude in Section~\ref{section:conclusion}.
\section{Proposed approach}
\label{section:method}
In this section, we will describe the {\it $DC^2$} algorithm for large scale kernel learning. {\it $DC^2$} consists of three 
steps: 1) data partition; 2) representation compression; 3) conquering and aggregation of subtasks. For concreteness of description, we will 
apply the {\it $DC^2$} algorithm to {\it rpfCluster} \cite{rpfClusterArXiv2019}, a clustering algorithm based on a data-driven kernel 
learned by random projection forests (rpForests) \cite{rpForestsArXiv2018}; the resulting algorithm is termed {\it rpfCluster\textsuperscript{+}}. 
We will describe the {\it $DC^2$}  algorithm and {\it rpfCluster\textsuperscript{+}} in the rest of this section.
\subsection{The {\it $DC^2$} algorithm}
\label{section:DCC}
\begin{figure}[h]
\centering
\begin{center}
\hspace{0cm}
\includegraphics[scale=0.32,clip]{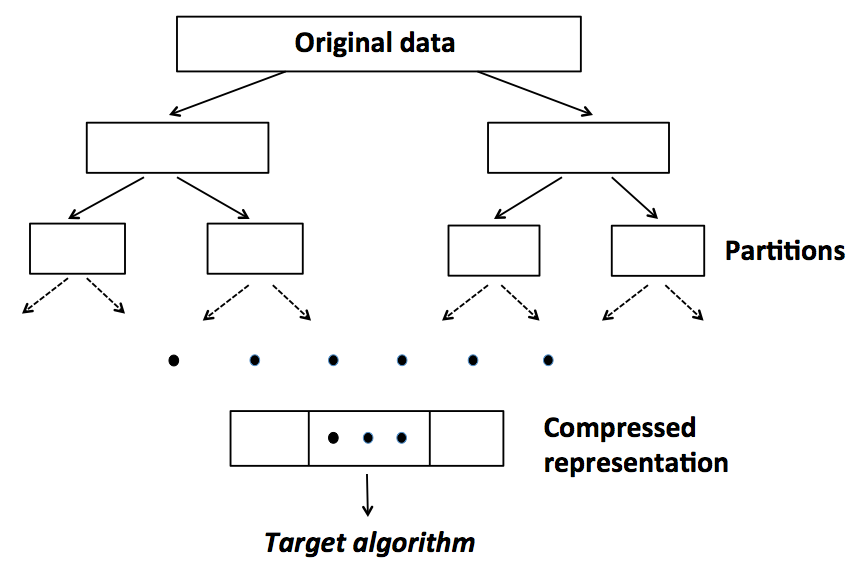}
\end{center}
\caption{\it Illustration of the {\it $DC^2$} algorithm. } 
\label{figure:archDcc}
\end{figure}

We divide the data by recursive random projections. We first project all the data onto a randomly generated direction, then divide 
the data into two halves by the median of the projections (or by a randomly chosen split point). If more than two partitions are required, 
then we continue on each of the two halves recursively until reaching the number of partitions. Compared to random sampling based 
partition, our approach incorporates the geometry in the data thus better or more robust results are expected. Later in our experiment, 
we will demonstrate that this geometry-awareness is empirically desirable. As we use recursive random partitions to split the data, 
representation compression will be very handy. On each data partition, we simply continue the random partition of the data until the size 
of the leaf node is ``small'' enough. We then compress each leaf node to its centroid or a randomly picked point within the leaf node. We 
refer to the compressed point as the signature of the associated leaf node. Thus each data partition in {\it $DC^2$} will 
produce a number of signature points with each corresponding to a leaf node in that partition.
\\
\\
For the aggregation of subtasks, we simply collect signature points across all data partitions. {\it The collection of all signature 
points}, denoted by $S$, will then be fed to some kernel learning algorithm (i.e., {\it rpfCluster} in the present work).
Figure~\ref{figure:archDcc} illustrates the architecture of the {\it $DC^2$} algorithm.
\\
\\
As we produce data partitions via recursive random projections and generate compressed representations also by recursive random 
projections on each data partitions, our approach can be viewed as unifying the division (of the data) and conquering (of the 
subproblems) with recursive random projections (i.e., rpTrees \cite{DasguptaFreund2008, rpForestsArXiv2018}). The top few levels
of rpTrees are used to divide the data, and further levels (i.e., subtrees) are used to generate compressed representation for each 
data partition. The level to cut the full rpTrees for subtrees depends on the number of partitions to use in divide-and-conquer. For 
example, if we aim at 2 partitions then cut at the 1\textsuperscript{st} level (the root node, or the original data, has a level 0) with each 
of the two child nodes becoming the root of a subtree. 4 partitions can 
be obtained by cutting at the 2\textsuperscript{nd} level, and so on. Each subtree becomes one partition of the data. The advantage of splitting 
the data recursively, instead of partitioning all at once along a single random direction, is to avoid thin and long slices of data 
which will potentially harm the performance of the subsequent clustering \cite{rpForestsArXiv2018}. 
\\
\\
Now we can give an algorithmic description of the {\it $DC^2$} algorithm. Let $V$ denote the set of all points. Let $n_p$ denote the 
predefined number of data partitions (e.g., same as the number of cores on the machine), and $n_s$ be the node size below
which rpTrees will not grow further. As {\it $DC^2$} uses rpTrees, we also include a description of rpTrees. The algorithm for {\it $DC^2$} 
and rpTrees are termed as Algorithm~\ref{algorithm:dcc} and Algorithm~\ref{algorithm:rpTree}, respectively.
\begin{algorithm}[h]
\caption{\it~~rpTree(D)}
\label{algorithm:rpTree}
\begin{algorithmic}[1]
\STATE Let $D$ be the root node of tree $t$; 
\STATE Initialize the working queue $\mathcal{W} \leftarrow \{D\}$; 
\WHILE {$\mathcal{W}$ is not empty}
	\STATE Take node $W$ from $\mathcal{W}$;
	\STATE If $|W| < n_s$, then skip to next round of the loop; 
    	\STATE Split node $W$ by random projection into $W= W_L \cup W_R$;
	\STATE Add $W_L, W_R$ to queue $\mathcal{W}$ and also tree $t$; 
\ENDWHILE
\STATE return($t$); 
\end{algorithmic}
\end{algorithm} 
\begin{algorithm}[h]
\caption{\it~~{\it $DC^2$}(V)}
\label{algorithm:dcc}
\begin{algorithmic}[1]
\STATE Initialize $S \gets \emptyset, \mathcal{W} \leftarrow \{V\}$; 
\FOR {$i=1$ to $n_p$}
	\STATE Take the largest node $W$ from $\mathcal{W}$;
    	\STATE Split node $W$ by random projection into $W= W_L \cup W_R$;
	\STATE Add $W_L, W_R$ to queue $\mathcal{W}$; 
\ENDFOR
\FOR {$i=1$ to $n_p$}
\STATE Take node $W$ from queue $\mathcal{W}$; 
\STATE Grow random projection tree $t_i \gets rpTree(W)$; 
\STATE Let $S_i$ be the set of leaf node signatures for tree $t_i$; 
\STATE Update $S \gets S \cup S_i$; 
\ENDFOR
\STATE return($S$); 
\end{algorithmic}
\end{algorithm} 
\subsection{{\it rpfCluster}}
{\it rpfCluster} \cite{rpfClusterArXiv2019} is a clustering algorithm based on the learning of the {\it rpf-kernel} via {\it rpForests} 
\cite{rpForestsArXiv2018}. {\it rpForests} is an ensemble of random projection trees (rpTrees) \cite{DasguptaFreund2008} with 
the possibility of projection selection during tree growth. Instead of splitting the nodes along coordinate-aligning axes such as the
popular kd-tree \cite{Bentley1975}, rpTrees recursively splits the tree along randomly chosen directions. {\it rpForests} combines 
the power of ensemble methods \cite{Bagging,RF,Adaboost,CF} and the flexibility of trees. {\it rpForests} is computationally efficient 
with a log-linear average complexity for growth and $O(\log(n))$ for search. As the tree partitions the data space recursively, data points 
falling in the same tree leaf node would be close to each other or ``similar''. This property is leveraged for the construction of the 
rpf-kernel. Additionally, as individual trees are rpTrees, {\it rpForests} can adapt to the geometry of the data and readily overcomes the curse 
of dimensionality \cite{DasguptaFreund2008}.
\\
\\
The rpf-kernel is constructed by averaging the incidence matrices induced by trees in {\it rpForests}. On each tree, an incidence 
matrix is created with its $(i,j)$ position being 1 if the i\textsuperscript{th} and j\textsuperscript{th} points lie in the same leaf node and 0 otherwise. The rpf-kernel 
is further transformed by $exp(S/\beta)$ for some properly chosen bandwidth $\beta$ to reflect the correct scale at which the data 
are clustered (just like the Gaussian kernel). {\it rpfCluster} then works by applying spectral clustering 
to the rpf-kernel. The cluster membership from spectral clustering is then used, along with a correspondence between the signature 
point and all points in the same leaf node in {\it $DC^2$}, to derive the cluster membership for all the original data 
points. Let $T$ denote the number of trees in {\it rpForests}. {\it rpfCluster} is described as Algorithm~\ref{algorithm:rpfCluster}.
\begin{algorithm}
\caption{\it~~rpfCluster($ \mathcal{S}$)} 
\label{algorithm:rpfCluster}
\begin{algorithmic}[1]
\STATE Initialize a similarity matrix $K \gets \bf{0}$; 
\FOR {$i=1$ to $T$}
\STATE Grow random projection tree $t_i \gets rpTree(S)$; 
	\FOR {each leaf node $\mathcal{N}\in t_i$}
	\STATE Increase the similarity count for each entry in $K[\mathcal{N},\mathcal{N}]$; 
	\ENDFOR
\ENDFOR
\STATE Average $K$ by $K \gets K/T$; 
\STATE $K \gets exp(K/\beta)$ for some bandwidth $\beta$;
\STATE Apply spectral clustering to $K$; 
\STATE Populate spectral clustering membership to all data points;
\end{algorithmic}
\end{algorithm} 
\\
For details about spectral clustering, the reader can refer to \cite{Ncut, NgJordan2002, Luxburg2007}.
\section{Theoretical analysis when {\it $DC^2$} is applied to spectral clustering}
\label{section:theory}
In applying the {\it $DC^2$} algorithm, a data point is replaced by a signature point (indeed many points falling into the same leaf node 
of the rpTrees would be replaced by the same signature point). Due to their discrepancy, there will be an 
approximation error in the resulting kernel matrix and the subsequent kernel learning. Error bounds can be derived for several 
kernel learning methods in \cite{CortesMohriTalwalkar2010}, including kernel ridge regression, support vector machines, 
graph Laplacian regularization algorithms. We focus specifically here in deriving an error bound when spectral clustering 
is applied to the kernel learned on the set of signatures generated by {\it $DC^2$}. The discrepancy is modeled as data perturbation
in our analysis.
\\
\\
We treat data perturbation as adding a noise component $\epsilon$ to data $X$ \cite{mynips2008, YanHuangJordan2009tech}
\begin{equation}
\label{eq:noise}
\tilde{X}=X+\epsilon.
\end{equation}
Assume $\epsilon$ is symmetric about $0$ with bounded 
support, and let $\epsilon$ have standard deviation $\sigma_{\epsilon}$ that is small compared to $\sigma$, the standard 
deviation for the distribution of $X$. Note that here we assume that $X\in \mathbb{R}$; this is for simplicity of discussion,
and the extension to $\mathbb{R}^d$ is straightforward.
\\
\\
To prepare for the perturbation 
analysis, let us introduce some notations. Let $X_1, ..., X_N$ be the given data. Let $\mathcal{K}(.,.)$ be the similarity kernel of interest. 
Denote $a_{ij}=\mathcal{K}(X_i, X_j)$ and 
let $A=(a_{i,j})_{i,j=1}^N$ be the similarity matrix. Let $\mathcal{L}$ be the graph Laplacian of $A$, i.e., 
\begin{equation}
\label{eq:defLaplacian}
\mathcal{L}=D^{-\frac{1}{2}}(D-A)D^{-\frac{1}{2}},
\end{equation}
where $D=diag(d_{1},...,d_{N})$ with $d_i=\sum_{j=1}^N a_{ij}, i=1,...,N$. We will use $\sim$ to denote quantities due to 
perturbation, e.g., $\tilde{a}_{ij}=\mathcal{K} (\tilde{X}_i, \tilde{X}_j)$.
Our analysis is based on an end-to-end error bound w.r.t. the distortion to the Laplacian matrix \cite{mynips2008}, and a 
perturbation bound to the Laplacian matrix due to data perturbation \cite{YanHuangJordan2009tech}. 
The main result of our perturbation analysis is stated as Theorem~\ref{thm:pertBound}.
\begin{theorem}
\label{thm:pertBound}
Suppose $X_1,...,X_N$ is a sample of bounded support such that $\inf_{1\leq i \leq N} d_i /N>\delta_0$ holds 
in probability for some constant $\delta_0>0$. Assume the similarity kernel $\mathcal{K}(.,.)$ is uniformly bounded and further 
there exists universal constant $K$ s.t. $|\mathcal{K}(\tilde{X}_1, \tilde{X}_2) - \mathcal{K}(X_1, X_2) |
\leq K \left(\sum_{i=1}^2 ||\tilde{X}_i - X_i||^2 \right)$ where $||.||$ is the Euclidean norm. Assume 
the data perturbation $\epsilon$ is symmetric about $0$ with bounded support. Then under suitable technical conditions, the
loss $\rho$ in clustering accuracy of a spectral bi-partitioning algorithm
due to $DC^2$ satisfies
\begin{eqnarray*}
\rho \leq ||\tilde{\mathcal{L}}-\mathcal{L}||_F^2 \leq_{p}
C \sigma_{\epsilon}^2 + C' \sigma_{\epsilon}^4,
\end{eqnarray*}
where $||.||_F$ indicates the Forbenius norm, for some constants $C$ and $C'$, as $N \rightarrow \infty$.
\end{theorem}
\noindent
Theorem~\ref{thm:pertBound} states that the loss in clustering accuracy due to data perturbation is bounded by the 
second and fourth moments of the data approximation error. According to \cite{DasguptaFreund2008}, the radius of 
tree leaves (i.e., the distance between a point and the node centroid) in rpTrees vanishes. Thus the mean data approximation 
error of the full data by the signature points vanishes and so does the perturbation bound and the loss in clustering accuracy 
when the number of signature points increases. 
\section{Related work}
\label{section:related}
There are several lines of work that are related to ours. This includes many work that use the divide-and-conquer principle to 
tame large scale problems. An influential line of work is {\it Bag of Little Bootstrap} \cite{BagLittleBootstrap}, a big data version 
of Bootstrap \cite{Efron1979}. The idea is to take many very ``thin" subsamples to be distributed to many computer nodes, and 
then results from those individual subsamples are aggregated. \cite{ChengShang2015} considers smoothing spline and explored 
the tradeoff between computational efficiency and statistical optimality of the {\it Divide-and-Conquer} methods in a distributed 
environment. \cite{ChenXie2014} studied penalized regression for data too big to fit in the memory by 
working on subsamples of the data and then aggregating the resulting models. Also, \cite{VolgushevChaoCheng2019} 
learns a quantile function, 
\cite{ZhangDuchiWainwright2015} 
studies ridge regression, \cite{BatteyFan2015} considers the general distributed estimation and inference, \cite{LeeLiuSunTaylor2017} 
learns a Lasso-type linear model at individual sites and then aggregate to de-bias, \cite{RichtarikTakac2016} explores coordinate descent 
for distributed data, \cite{RosenblattNadler2016} studies the optimality of averaging in distributed computing. Such work are different 
from ours in that they all {\it average} results obtained on random subsamples,
while we {\it assemble} results from geometry-aware data partitions with {\it $DC^2$} algorithm.
\\
\\
Another line of closely related work are those under the term ``learning over inherently distributed data'' \cite{distSpectArXiv2019, distStatArXiv2019}. 
Instead of dividing the data, these work deal with situations where the data are already distributed, i.e., stored at a number of distributed 
machines as a result of business operation or diverse data collection channels. Computations are performed on the local data on the 
machine that stores the data, and then local signatures are sent to a central server for aggregation. Our work almost works in a reverse 
way, which splits and distributes data to multiple computer cores or machines for parallel computation. Of course, there are also work that 
use the idea of data compression for approximate large scale computation. This includes fast approximate spectral clustering 
\cite{YanHuangJordan2009tech}, landmark based spectral clustering \cite{Chen2011LargeSS}, and also spectral clustering by the 
Nystr\"om method \cite{williams-seeger-nystrom, NystromSpectral}. While all use data compression, our method starts with divide-and-conquer 
and further employs parallel computation. 
\section{Experiments}
\label{section:evaluation}
Our experiments consists of three parts. We first evaluate the mean squared errors (MSE) in approximating the full data by the collection 
of signature points when the data is partitioned by random sampling 
versus by random projection. Then we evaluate {\it rpfCluster\textsuperscript{+}} by comparing it to competing algorithms, including K-means clustering 
\cite{HartiganWong1979} as the baseline, and two algorithms, KASP and RASP, for large scale spectral clustering 
\cite{YanHuangJordan2009tech}. In the third part, we evaluate the performance of {\it rpfCluster\textsuperscript{+}} under different data partition 
schemes and different number of partitions. We start by describing the data, the performance metrics and competing 
methods.
\begin{table}[tbp]
\begin{center}
\setlength{\extrarowheight}{1pt}
\begin{tabular}{c|crc}
    \hline
\textbf{Data set}     & \textbf{\# Features}  &\textbf{\# instances}  &\textbf{\# classes} \\
    \hline
Connect-4   & 42       &67,557             &3\\
USCI           & 37       &285,779            &2\\
Cover Type	&54		&568,772				&5\\
HT Sensor   &11        &928,991             &3\\
Poker Hand  & 10       &1,000,000         &3\\
Gas Sensor  &18		& 8,386,765		 &2\\
\hline
\end{tabular}
\end{center}
\caption{\it A summary of UC Irvine data used in the experiments. } \label{tbl:data}
\end{table}
\\
\\
We use 6 benchmark datasets from the UC Irvine Machine Learning Repository \cite{UCI}, including the Connect-4, USCI (US Census Income), 
Cover type, HT Sensor, Poker Hand, and the Gas Sensor data. Table~\ref{tbl:data} is a summary of the datasets. For Connect-4, USCI, and 
Poker Hand data, we follow procedures described in \cite{YanHuangJordan2009tech} to preprocess the data. The original {\it USCI} data 
has 299,285 instances with 41 features. We exclude features \#26, \#27, \#28 and \#30, due to too many missing values, and then remove 
all instances with missing values. This leaves 285,799 instances on $37$ features, with all categorical variables converted to integers. The 
original {\it Cover Type} data has 581,012 instances. We excluded the two small classes (i.e., 4 and 5) for fast evaluation of accuracy 
(otherwise all 7! permutations need to be evaluated, but that is not the focus of the present work), and this leaves 568,772 instances; 
we also standardized each of the first 10 features to have a mean 0 and variance 1. The original {\it Poker Hand} data is highly unbalanced, 
with $6$ small classes containing less than $1\%$ of the data. Merging small classes gives $3$ final classes with a class distribution of 
$50.12\%$, $42.25\%$ and $7.63\%$, respectively. The {\it Gas Sensor} data consists of two different gas mixtures: Ethylene mixed with 
CO, and Ethylene mixed with Methane. The two different gas mixtures form two classes in the data. The Connect-4, the USCI, and the 
Gas Sensor data are standardized on all features.
\\
\\
The performance is assessed by clustering accuracy and the computation time. The use of clustering accuracy aligns closely to 
the ultimate goal of clustering---assigning data points to proper groups. In contrast, many other performance metrics are often a 
surrogate of this due to the lack of true labels. However, as we are evaluating clustering algorithms, we have 
the freedom to use data with true labels. 
\\
\\
\textbf{Definition.} Let $\mathcal{L}=\{1,2,...,l\}$ be the label set. Let $h(.)$ and $\hat{h}(.)$ be the true label and the label 
obtained by a clustering algorithm, respectively. The {\it clustering accuracy} is defined as
\begin{equation}
\label{clusterAccuracy} \rho_c(\hat{h})=\max_{\tau \in \Pi_{\mathcal{L}}}
\left\{\frac{1}{n}\sum_{i=1}^n \mathbb{I}\{\tau
\left(h(X_i)\right)=\hat{h}(X_i)\}\right\},
\end{equation}
where $\mathbb{I}$ is the indicator function and $\Pi_{\mathcal{L}}$ is the set of all permutations on the label set $\mathcal{L}$. 
It measures the fraction of labels by a clustering algorithm that agree with the true labels that come with the dataset
up to a permutation of the true labels. This is a natural extension of the classification accuracy (under 0-1 loss) 
and has been used by many work in clustering \cite{XingNgJordanRussell2002,MeilaShortreed2005, YanHuangJordan2009tech}. 
\subsection{Competing methods and parameters}
\label{section:competitors}
We compare {\it rpfCluster\textsuperscript{+}} to three other clustering algorithms---K-means clustering, and two variants of fast approximate 
spectral clustering algorithms, KASP and RASP \cite{YanHuangJordan2009tech}. {\it Note that here our goal is not to show which 
algorithm outperforms others, but to demonstrate that our {\it $DC^2$} algorithm can significantly speed up large scale kernel 
learning (with a data-driven similarity kernel)}. 
\\
\\
K-means clustering \cite{lloyd1982} is one of the most widely used clustering algorithms. It  starts with randomly generated 
cluster centroids, and then alternates between two steps: 1) assign data points to the closest centroid; and 2) recalculate 
the cluster centroid for each cluster, until the change to the within-cluster sum of squares is small enough. The R package 
{\it kmeans()} is used with the ``Hartigan-Wong" \cite{HartiganWong1979} initialization, and the maximum number of iterations 
and the number of restarts are set to be $(200,20)$. 
\\
\\
KASP and RASP are among the fastest algorithms for spectral clustering. Both are based on the idea of shifting expensive 
spectral clustering to a small amount of structure-preserving data signatures. While KASP obtains data signatures via K-means 
clustering, RASP grows rpTrees. The data compression ratio is chosen such that 
the signature set has about 500-1000 points. The bandwidth parameter for the Gaussian kernel varies with a step size of 0.1 in 
the range [0.1,1] and 1 in (1,200].
\\
\\
For {\it $DC^2$}, the data compression ratio is such that the collection of signature points has a size about 1000. For {\it rpfCluster}, 
the number of trees is fixed at 800, the node splitting constant parameter $n_s$ is fixed at $30$, and the step size for 
the search of bandwidth $\beta$ is 1 in the range [10, 80]. All results in our experiments are averaged over 100 runs.
\subsection{Partition by random sampling or projection}
\label{section:mses}
We advocate the use of recursive random projections for data partitions in the {\it $DC^2$} algorithm. In Section~\ref{section:DCC}, we 
argue that this would lead to more robust algorithms for large scale kernel learning, and show that the loss in accuracy due to 
data reduction in {\it $DC^2$} is upper bounded by the approximation error of the full data by the collection of signature points. In 
this section, we will empirically demonstrate that partition by random projections tends to have a smaller mean squared approximation 
error. When the difference becomes significant (e.g., the ratio of MSEs is larger than 1.5), it will translate to a 
noticeable difference in clustering accuracy (c.f. Section~\ref{section:expPartitions}).
\begin{figure}[htbp]
\centering
\begin{center} 
\hspace{0cm}
\includegraphics[scale=0.56,clip]{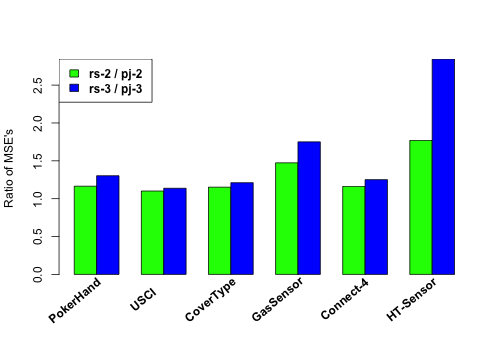}
\end{center}
\abovecaptionskip=-10pt
\caption{\it Ratio of MSEs in {\it $DC^2$} when dividing the data by random sampling (indicated by ``rs") 
and random projection (indicated by ``pj"). The number of data partitions are indicated in the legend.} 
\label{figure:comp2Vars}
\end{figure}
\\
\\
Figure~\ref{figure:comp2Vars} shows the ratio of MSEs due to the {\it $DC^2$} algorithm when dividing the data by random 
sampling and recursive random projections. It can be seen that, in all cases, the MSE by random sampling 
is larger than that by random projections. For two data sets, HT sensor and the Gas sensor, the ratios are larger than 1.5. Later in 
Section~\ref{section:expPartitions}, we will see that random sampling leads to a remarkably more loss in clustering accuracy on these 
two datasets. This gives support to our choice of recursive random projections for data partitions in the {\it $DC^2$} algorithm.
\subsection{Comparison of different clustering algorithms}
\label{section:expComps}
\begin{figure}[htbp]
\centering
\begin{center}
\hspace{0cm}
\includegraphics[scale=0.54,clip]{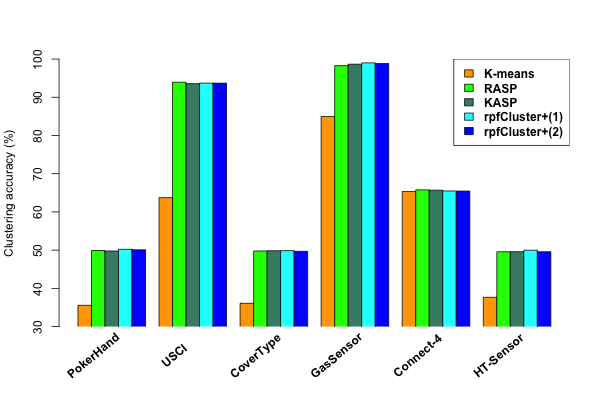}
\end{center}
\abovecaptionskip=-10pt
\caption{\it Clustering accuracy under K-means clustering, RASP, KASP, and {\it rpfCluster\textsuperscript{+}} (the numbers in the parenthesis indicate 
the number of partitions, and the partitions are obtained by random projections).} 
\label{figure:comp1AccM}
\end{figure}
\begin{figure}[htbp]
\centering
\begin{center}
\hspace{0cm}
\includegraphics[scale=0.54,clip]{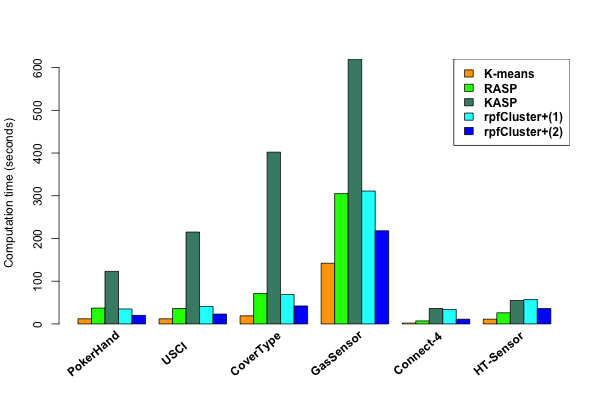}
\end{center}
\abovecaptionskip=-10pt
\caption{\it Clustering time under K-means clustering, RASP, KASP, and {\it rpfCluster\textsuperscript{+}} (the numbers in the parenthesis indicate 
the number of partitions, and the partitions are obtained by random projections).} 
\label{figure:comp1TM}
\end{figure}
\noindent
We compare the performance of {\it rpfCluster\textsuperscript{+}} to three competing algorithms, K-means clustering, KASP and RASP, on both
clustering accuracy and computation time. These are shown in Figure~\ref{figure:comp1AccM} and Figure~\ref{figure:comp1TM}, 
respectively. It can be seen that the clustering accuracy of K-means clustering is much lower than the other three algorithms on all
but one dataset, while that of the other three are quite similar. For computation time, K-means clustering is the fastest on all the 
datasets while KASP is the slowest on almost all the data. The computation time by {\it rpfCluster\textsuperscript{+}} (with one data partition) is close to 
that of RASP on all the data, and by running over two data partitions, {\it rpfCluster\textsuperscript{+}} becomes faster than RASP, and is close to 
K-means clustering on most of the datasets.
\subsection{Performance under varying number of partitions}
\label{section:expPartitions}
\begin{figure}[htbp]
\centering
\begin{center}
\hspace{0cm}
\includegraphics[scale=0.54,clip]{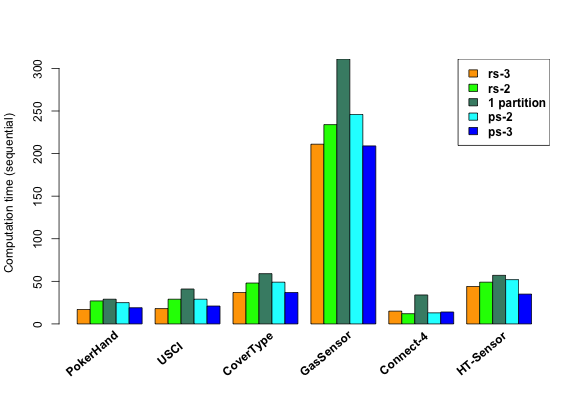}
\end{center}
\abovecaptionskip=-10pt
\caption{\it Clustering time under different schemes of dividing the data and with different partitions for rpfCluster\textsuperscript{+} running in
sequential mode. The number in the legend indicates the number of partitions.} 
\label{figure:comp2Tseq}
\end{figure}

We also evaluate the performance of {\it rpfCluster\textsuperscript{+}} under different schemes of data partitions, random sampling and random 
projection, with 2 or 3 partitions. 
\begin{figure}[htbp]
\centering
\begin{center}
\hspace{0cm}
\includegraphics[scale=0.54,clip]{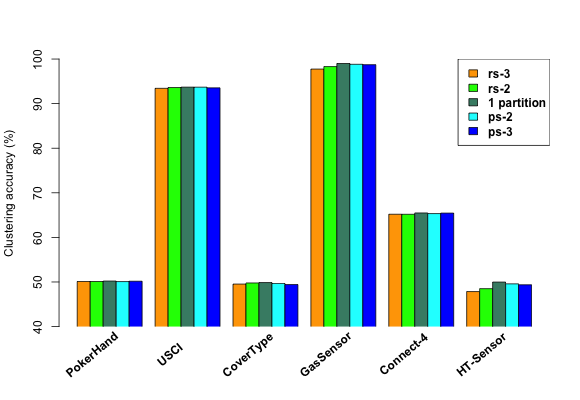}
\end{center}
\abovecaptionskip=-10pt
\caption{\it Clustering accuracy under different schemes of dividing the data and with different partitions for rpfCluster\textsuperscript{+}.
The number in the legend indicates the number of partitions.} 
\label{figure:comp2Acc}
\end{figure}
\begin{figure}[htbp]
\centering
\begin{center}
\hspace{0cm}
\includegraphics[scale=0.54,clip]{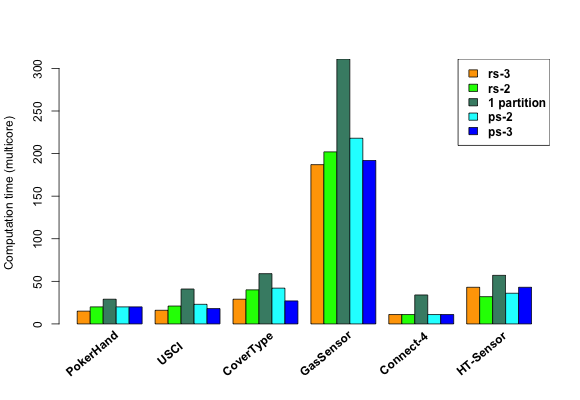}
\end{center}
\abovecaptionskip=-10pt
\caption{\it Clustering time under different schemes of dividing the data and with different partitions for rpfCluster\textsuperscript{+} running in
milticore mode.
The number in the legend indicates the number of partitions.} 
\label{figure:comp2TM}
\end{figure}
\noindent
Figure~\ref{figure:comp2Acc} and Figure~\ref{figure:comp2TM} show the clustering accuracy 
and computation time, respectively. In general, more data partitions leads to a shorter computation time; the gain in computational 
efficiency starts diminishing when more partitions are applied on smaller data. It is worth noting is that, on two datasets, the Gas 
sensor data and the HT-sensor data, the decrease in accuracy with more data partitions becomes noticeable when data partitions 
are obtained by random sampling while the loss of accuracy by random projections remains negligible. We attribute this to the much 
higher MSEs by random sampling, as discussed in Section~\ref{section:mses}. 
\\
\\
An additional experiment is conducted when {\it rpfCluster\textsuperscript{+}} is running in sequential mode, that is, assume there is only one core or a 
single machine in the cluster. Figure~\ref{figure:comp2Tseq} shows that there is a potential advantage in computational efficiency
to apply the {\it $DC^2$} algorithm to large scale problems, even if there is no parallel infrastructure (due possibly to the non-linearity
of the underlying kernel learning algorithm).  
\section{Conclusions}
\label{section:conclusion}
We have proposed an effective algorithm, {\it $DC^2$}, for large scale kernel learning. {\it $DC^2$} applies divide-and-conquer with a further distortion 
minimizing representation compression on the data, and achieves the efficiency of the conventional sampling based approach for large scale 
computation with a potential of parallel computation on multicore or clustered computers. With {\it $DC^2$}, the partition and the subsequent 
representation compression of data are implemented under a unified operation---recursive random projections. We advocate the use of 
recursive random projections for dividing the data in divide-and-conquer, which has the advantage of smaller MSEs compared to the 
conventional approach of random sampling. On a random projection forests based clustering algorithm, we demonstrated the power and 
efficiency of {\it $DC^2$} algorithm (resulting algorithm termed as {\it rpfCluster\textsuperscript{+}}). {\it rpfCluster\textsuperscript{+}} achieves 
a similar level of clustering accuracy 
as KASP and RASP, some of the fastest approximate spectral clustering algorithms, and has a running time close to that of K-means clustering. 
Theoretical analysis is carried out on {\it $DC^2$} when the resulting data signatures are used as input to spectral clustering, and we show that 
the loss in accuracy due to data reduction is upper bounded by the data approximation error which would vanish with recursive random 
projections. Due to the easy implementation and flexibility of {\it $DC^2$}, we expect it to be applicable to general large scale learning and 
inference problems.

\end{document}